# Automatic Generation of Highlights for Academic Paper Via Prompt-based Learning

Yi Xiang, Chengzhi Zhang *, Heng Zhang

Department of Information Management, Nanjing University of Science and Technology, Nanjing, China,

**Abstract:**

**Purpose -** The highlight of an academic paper is a condensed summary of the author's critical work, enabling readers to understand the paper's focus quickly. However, many journals currently do not provide highlights for papers. Some scholars have employed supervised learning methods to extract highlights from academic papers to address this issue. One challenge in this research is the requirement for a large amount of training data.

**Design/methodology/approach -** In this study, we investigated the effectiveness of using Prompt-based learning to generate highlights. We constructed specific prompt templates based on the task requirements, which we then filled with paper abstracts in prompt and input into language models to generate the highlights. For the language models, we used pre-trained models that can perform inference locally, such as GPT-2 and T5, as well as the ChatGPT model accessed through an API.

**Findings -** By validating the model's generation performance on three datasets, we found that the performance of the ChatGPT model, when coupled with the prompt templates, was comparable to past supervised learning methods, which were achieved without any training samples. When we added a small number of training samples to the prompt templates, the model's performance on two datasets significantly surpassed state-of-the-art approaches. We also analyzed the impact of the prompt template content on the model's performance. The results showed that although the ChatGPT model has powerful language modeling capabilities from training on massive-scale corpora, its performance on specific tasks is heavily influenced by the information provided in the prompt templates. Finally, we compared the highlights written by the authors with those generated by the model through specific examples. From the content perspective, the method used in this study generated highlights of relatively high quality.

**Originality/value -** This study contributes to the research on automatic highlight extraction. It is the first to apply the prompt learning approach to highlight generation research, utilizing multiple mainstream pre-trained language models and the currently popular ChatGPT model for text generation. Notably, the method does not require training on domain-specific corpora, making it more widely applicable. The proposed method can automatically generate highlight content for many papers that initially lacked such information. This, in turn, can provide data support for subsequent text mining and bibliometric research based on highlight information.

* Corresponding author, Email: zhangcz@njust.edu.cn

## 1 Introduction

Academic articles are vital in mediating research achievements and scientific knowledge (Zhang C et al., 2023). They are essential resources for researchers to learn new techniques and stay updated with the latest advancements(Liu, 2023). However, the volume and diversity of academic literature are constantly growing (Cao et al., 2023), posing a significant challenge in retrieving and analyzing the contents of papers. Even experienced researchers may need help locating relevant works amidst massive literature. Additionally, longer article lengths make quick reading more difficult(Lu et al., 2017). The main issue lies in filtering the article's information and extracting its critical content. In this study, we propose a method to address the below problem by developing an automated system to generate highlight summaries for scholarly articles. Highlights are sentences in an academic article that effectively convey its main content (Collins et al., 2017), which are more concise than the abstract and facilitate quick reading by readers. Figure 1 illustrates how the abstract and highlights sections of an article differ in terms of content and presentation; by referring to highlights instead of reading a lengthy abstract, we can quickly grasp that the current research aims to enhance prompter sequence prediction performance using a combined approach of prompt learning and pre-trained language models. ScienceDirect also recommends that authors provide the paper's highlights as they improve search engine retrieval possibilities so that interested peers can notice these works.

Abstract

The promoter region, positioned proximal to the transcription start sites, exerts control over the initiation of gene transcription by modulating the interaction with RNA polymerase. Consequently, the accurate recognition of promoter regions represents a critical focus within the bioinformatics domain. Although some methods leveraging pre-trained language models (PLMs) for promoter prediction have been proposed, the full potential of such PLMs remains largely untapped. In this study, we introduce PLPMpro, a model that capitalizes on prompt-learning and the pre-trained language model to enhance the prediction of promoter sequences. PLPMpro effectively harnesses the prompt learning paradigm to fully exploit the inherent capacities of the PLM, resulting in substantial improvements in prediction performance. Experiment results unequivocally demonstrate the efficacy of prompt learning in bolstering the capabilities of the pre-trained model. Consequently, PLPMpro surpasses both typical pre-trained model-based methods for promoter prediction and typical deep learning methods. Furthermore, we conduct various experiments to meticulously scrutinize the effects of different prompt learning settings and different numbers of soft modules on the model performance. More importantly, the interpretation experiment reveals that the pre-trained model captures biological semantics. Collectively, this research imparts a novel perspective on the optimal utilization of PLMs for addressing biological problems.

PLPMpro: Enhancing promoter sequence prediction with prompt-learning based pre-trained language model

Highlights

- Propose a novel model combining prompt-learning and the pre-trained model to improve promoter sequence prediction.
- Experiments show the effectiveness of prompt learning in enhancing the capabilities of pre-trained models.
- Conduct experiments to scrutinize the effects of different prompt-learning settings on the model performance.
- Interpretation experiment reveals that the pre-trained model learns biological semantics.

**Figure 1. An example that revealed the difference between Abstract and Highlights**[1]

The advantages of highlights make it an efficient method for retrieving and reading articles. However, many journals do not provide corresponding highlights for their articles, except those included in ScienceDirect. To solve this problem, previous studies have crawled paper abstracts, highlight data from Elsevier as the training set, and then trained machine learning models to automatically extract or generate paper highlights from the abstracts(Collins et al., 2017; Cagliero & Quatra, 2020; Quatra & Cagliero, 2022). However, the issue with these methods is that they often require training multiple models for different domains, which increases the computational burden. Additionally, these methods typically require a sufficiently large training dataset to achieve good extraction performance, which means that researchers have had to spend a significant amount of time on dataset construction when they should have focused more on the design and optimization of the extraction model itself.

[1] https://www.sciencedirect.com/science/article/pii/S0010482523007254

Prompt-based learning is a valuable approach to address the limitations of training data size in supervised learning. It effectively enhances the performance of language models in low-resource scenarios. In past research, it has been expected to fine-tune models pre-trained on large-scale corpora to specific datasets, thereby applying the language modeling capabilities learned during the pre-training stage to the researchers' problems. Numerous studies have demonstrated the effectiveness of this approach (Liu & Zhao, 2022; Zheng et al., 2023). However, this method has a notable issue. There is often a significant inconsistency between the pre-trained model's pre-training objective and the model's fine-tuning objective. This mismatch can lead to a loss in model performance. Prompt learning addresses this issue by introducing prompt templates that transform the original input into a format the model has pre-trained to process. This allows the pre-trained language model to leverage its acquired capabilities better while maintaining good performance on downstream tasks with relatively less requirement for extensive training datasets.

Given the above research background, this study focuses on generating highlight content for papers automatically through prompt learning. The prompt learning method can reduce the model's demand for data volume and generate highlights for many papers without existing highlights, providing data support for subsequent large-scale bibliometric analysis research based on highlights. We selected five language models for the experiments and constructed nine specific prompt templates based on the characteristics of the highlight generation task. We then validated the method's effectiveness on three datasets, which provided abstracts of papers from the computer science, artificial intelligence, and biomedical domains, respectively. We filled the abstracts into the prompt templates as input to the language models, and without updating the model parameters, we took the model outputs as the generated highlights. We compared the generated highlights with those written by the paper authors to analyze the performance differences between different models and prompt templates. The research results show that the method used in this study can achieve performance comparable to past supervised models trained on large-scale datasets without any training samples. The model performance further improves When a few training samples are provided in the prompt templates. Additionally, we analyzed the impact of the prompt template content and the number of samples in the prompt templates on the model's generated results. Through specific examples, we demonstrated that the model-generated highlights can cover the research focuses emphasized by the paper authors well.

Our contributions are the following three folds:

(1) We applied the prompt-based learning approach to the task of highlight generation, surpassing the performance of previous supervised learning methods while reducing the demand for training data;

(2) We have thoroughly explored the impact of differences in prompt content on the performance of model generation;

(3) We analyzed the changes in performance before and after adding demonstrations, and through comparison, we found that using the n-gram overlap between test samples and selected training samples as a criterion for selecting demonstrations can lead to a more stable generation performance compared to random selection strategy;

This paper's data and source code are freely available at the GitHub website:

https://github.com/xiangyi-njust/Highlight-Generation.

# 2 Related work

This study focuses on using prompt learning methods to generate Highlights based on paper abstracts automatically. In the related work section, we first summarize the research on automatic Highlight extraction and generation. We then introduce the application of prompt learning methods in natural language processing.

## 2.1 Highlight Extraction and Generation

Highlight extraction is an automatic summarization task that first appears in the news corpus. Svore et al. (2007) believed that 3-4 relevant points for news articles would make it easier to read quickly than paragraph-style summaries. They use the extraction method and the neural network sorting algorithm to calculate the similarity between all sentences in CNN.com articles and the gold standard, then extract the highest similarity as the highlight. WoodSeed & Lapata (2010) used nearly 9,000 document-highlight pairs of CNN.com from 2007 to 2009 as a training and evaluation corpus and designed a joint content selection and compression model; the experimental results show that the highlight syntax and content output by this method are equivalent to those written by humans.

Collins et al. (2017) built the first dataset, CSPubSum, which contains academic article highlight information regarding academic article highlight extraction. They crawled the title, abstract (abstract), highlights and keywords, etc., regarded highlights as the gold standard for Text summarization tasks, and then built an automatic summarization model. Cagliero and La (2020) analyzed the difference between highlights and abstracts (abstract). They believed the automatic summarization system that generated high-quality abstracts needed more suitable for extracting highlights. They regarded the work of highlight extraction as a regression task, which is to find the sentence most similar to the highlight provided by the author from the full text of the article through feature extraction combined with machine learning methods such as Decision Tree. In addition, the author constructed two data sets, AIPubSum and BioPubSum, similar to Collins for model comparison. La & Cagliero (2022) introduced an attention mechanism to highlight extraction in their work in 2022. They still regard it as a regression task but use the Transformer architecture to encode semantic information better. The above works use the extraction method to obtain the article's highlight. Rehman et al. conducted several experiments to develop article highlights using a generated approach. They (Rehman et al., 2020) compared the performance of two networks, namely the pointer-generator network and the seq-to-seq network with attention mechanism; the former produced more meaningful highlight expressions. In subsequent research, Rehman enhanced the model's semantic representation of article text by integrating SciBert (Rehman et al., 2023a) and ELMO (Rehman et al., 2023b) pre-trained embeddings into the pointer-generator network, resulting in optimized generation performance.

## 2.2 The application of Prompt-based Learning in the NLP field

As a new research paradigm in the field of NLP, prompt-based learning has proven effective on many tasks. For the text classification task, Schick & Schütze (2021) used a semi-supervised method, *Pattern Exploiting Training* (PET), to solve text classification and natural language reasoning problems in low-resource scenarios. PET first reformulates the input text into a cloze form to help the language model better understand the given task and then fine-tunes the language model based on these phrases to assign soft labels to unlabeled instances and finally assign labeled ones for supervised learning. Experimental results show that this method can effectively improve model performance in low-resource scenarios; Han et al. (2021) demonstrated that manually formulated prompts are unsuitable for handling some classification tasks where finding suitable words to distinguish different categories is difficult. Therefore, they manually designed some sub-prompts, then applied logical rules to combine the sub-prompts into the final prompt and applied it to the fine-tuning process of the pre-trained model. They used this method on the complex multi-category classification task of relationship classification and significantly surpassed the existing SOTA on four relevant data sets. Song et al. (2023) employed prompt learning to address the event classification task with limited labeled samples. Initially, they automatically generated templates relevant to the specific task information. Subsequently, they utilized the Fisher information filtering method to select the most appropriate example samples for each query. These selected samples were then used to train the model.

Liu et al. (2022) proposed a named entity recognition method based on prompt-based learning for the information extraction task. They converted the NER problem into the form of QA and then automatically generated prompts suitable for the QA problem. The performance of this method in zero-shot and few-shot settings is better than some previous work; Kong et al. (2023) applied prompt in the keyword extraction process. They chose the T5 model for experiments. The encoder-side prompt is "*Book: [D]*" and fills *[D]* with the documents to be extracted; the decoder-side prompt is "*This book mainly talks about [C]*," and *[C]* corresponds to the candidate keywords output by the model, the final keyword set is determined through the candidate keyword probability and the position penalty introduced by the author. Experiments demonstrated that this method performs best on all the tested datasets.

The above work introduces prompts to reorganize the input for different tasks. Pre-trained language models like T5 and GPT2 generate outputs based on these prompts as context without updating model parameters (although fine-tuning can also be performed). This approach has shown excellent performance in zero-shot and few-shot learning scenarios. Previous studies often relied on supervised learning methods to solve the highlight generation task, where the amount of training data heavily influenced model performance. To reduce this dependency while maintaining high model performance and enabling quick application to other domains' data, this paper studies how prompt-based learning can be utilized for highlight generation tasks.

# 3 Methodology

The research task of this paper is to automatically generate highlights based on the abstract content of academic papers. We use a prompt learning approach to tackle this task, and the specific experimental procedure is shown in Figure 2. First, we use web crawling technology to obtain abstracts and highlight data from papers on the Elsevier website. Then, we design several prompt templates based on the characteristics of the current task, and the paper abstracts are filled into the prompt templates as the input for the subsequent generation model. Finally, five pre-trained language models are used as the generation methods in this experiment, with the prompt templates from the previous stage as the model input to generate the corresponding highlight content.

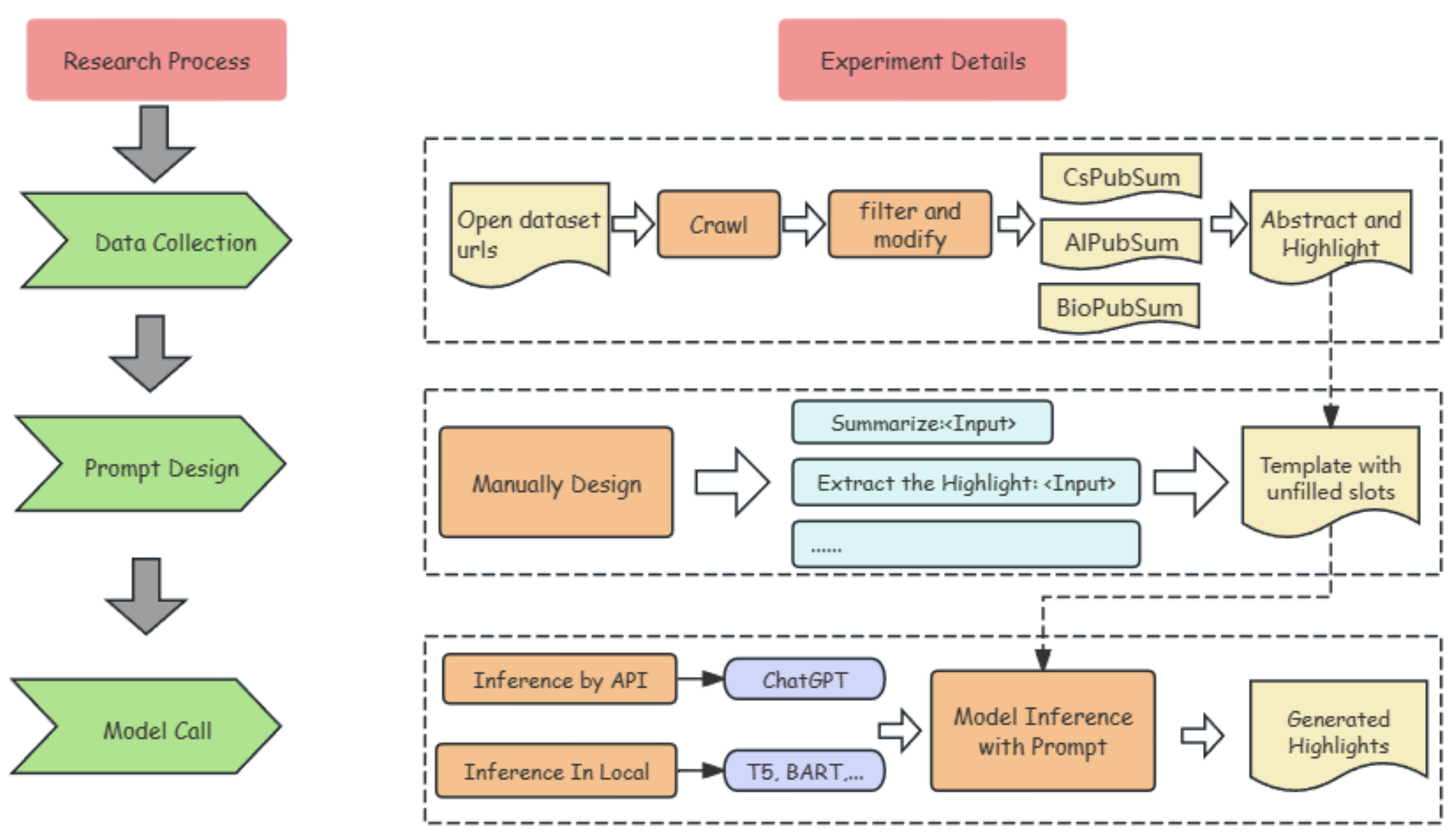

**Figure 2. The framework of this study**

## 3.1 Data Collection

The online versions of most papers in the journals published by Elsevier provide Highlights information. From the data acquisition perspective, we can crawl information of any domain and quantity for the experiments in this paper. However, to compare with the methods used in previous studies, this experiment selects three datasets that have been used in past research on Highlights extraction tasks: CSPubSum (Collins et al., 2017), AIPubSum and BIOPubSum (Cagliero & La, 2020), which cover paper abstracts and highlight content from the computer science, artificial intelligence, and biomedical domains, respectively. One issue that arose during the dataset acquisition process is that the authors of the three datasets mentioned above only provided the web links to the online versions of the papers. Therefore, we utilized the Octopus web crawling tool to crawl the abstracts and highlight the content of the papers based on the links provided in the datasets. We then processed the crawled data, including removing any records where the crawling resulted in empty content and adjusting the format of highlight content by eliminating redundant blank lines and line breaks.

Table 1 presents the descriptive statistics of the datasets. The data scales of the three datasets differ to some extent. The AIPubSum dataset has the smallest number of samples in both the training and test sets. The BioPubSum test set has a relatively large scale, with more than 2,000 samples. The CSPubSum dataset has the largest training set sample size. This has led to models in previous supervised highlight extraction studies performing better on the CSPubSum dataset than the other two. In comparison, an advantage of the method proposed in this paper is that we only need to include a small batch of training samples within the prompt templates to enable the model to learn more effectively (as shown in Section 3.2, we use at most six samples from the training set in the templates). This allows us to avoid the issue of poor extraction or generation performance due to small training set sizes. Additionally, we examined the average number of highlights per paper across the different datasets. We found that this value is similar between the datasets, with each paper providing around four highlights on average. This consistent finding provided a helpful reference point when designing the prompt templates for our subsequent experiments.

**Table 1 Descriptive Statistics of Datasets**

| Dataset | CSPubSum | AIPubSum | BioPubSum |
|---|---|---|---|
| training set | 10145 | 197 | 8056 |
| test set | 150 | 65 | 2685 |
| the average number of highlights in a single article | 4.23 | 4.18 | 3.99 |

## 3.2 Prompt Design

According to whether there is a training process, the algorithms based on Prompt-based Learning can be divided into two types: 1) The original input is transformed into a new input through a prompt template, and the new input-output pair is formed with the original label to train the model; 2) The original input is transformed into a new input through a prompt template, and the model generates directly based on the new input without parameter updates. The latter is the approach used in the research presented in this paper.

The prompt template is a text with placeholders for the content to be filled in, explicitly informing the model about the task. The research presented in this paper aims to generate Highlights based on the paper's abstract. A simple template used in this work is: "Summarize the highlight: [X][Y]," where [X] represents the placeholder for the paper's abstract (the input), and [Y] corresponds to the model's output. The choice of different prompt templates can significantly impact the model's performance for a specific research task. In this paper, we have provided nine prompt templates, as shown in Table 2. These templates vary in length and the level of detail in the content. In Section 4.3, we compare the performance differences of the models when using these different prompt templates.

**Table 2. The prompts applied in the experiment**

| No. | Template format |
|---|---|
| 1 | Summarize: [X][Y] |
| 2 | Summarize the highlight: [X][Y] |
| 3 | Summarize the main work: [X][Y] |
| 4 | Extract the highlight: [X][Y] |
| 5 | Extract the main work: [X][Y] |
| 6 | Refine 4~5 highlights of the article from the abstract: [X][Y] |
| 7 | Refine the Highlight of the article from the abstract: [X][Y] |
| 8 | Refine 4~5 innovative points of the scientific article abstract provided below in bullet point format without additional explanation: [X][Y] |
| 9 | Refine the innovation points of the scientific article abstract provided below in bullet-point format without additional descriptions: [X][Y] |

It is worth noting that with the emergence of models like GPT-3 and ChatGPT, a new prompt-based learning approach called In-context Learning (Brown et al., 2020) has demonstrated impressive performance across various tasks. In-context Learning involves including a certain number of input-output examples within the prompt, enabling the model to understand the current task at hand better. As illustrated in Figure 3, the original prompt template corresponds to the Task Definition section. In the Demonstration section, we select several abstracts and their corresponding Highlights from the training set as examples. This enables the model to learn better which information in the abstract is relevant to the Highlight and standardizes the output format. The Inference input refers to the abstract text for which a Highlight needs to be generated. These three components - Task Definition, Demonstration, and Inference input - collectively constitute the new Prompt. Given that the ChatGPT model was used in this experiment, we also considered this approach. One key question is how many data instances should be provided in the Demonstration section. The version of ChatGPT used in our experiment, gpt-3.5-turbo-0613, supports a maximum input of 4096 tokens. We used the tiktoken tool to analyze the average token length of the abstracts and Highlights in the original dataset, as shown in Figure 4. The results indicate that most instances have fewer than 600 tokens. Therefore, we set the number of Demonstration examples to 6. (After deducting six demonstrations from 4096, there are approximately 500 remaining for our test sample input).

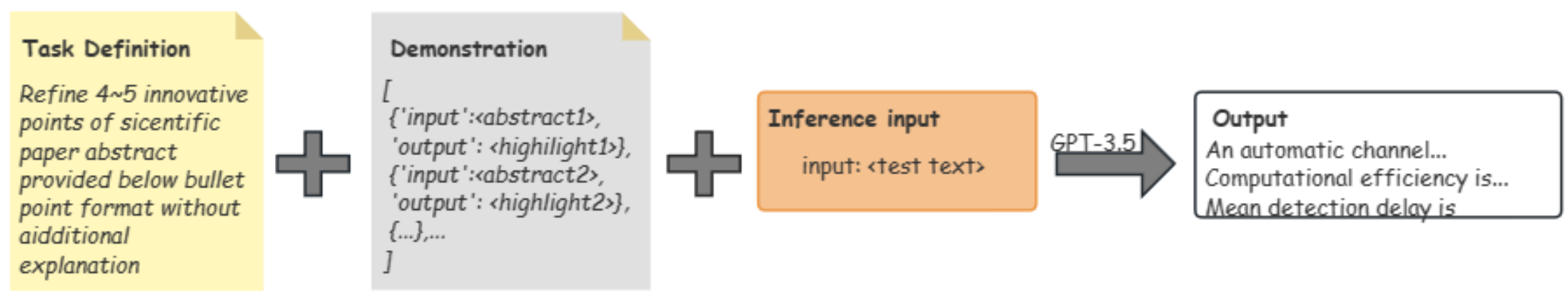

**Figure 3 The prompt format after adding demonstrations**

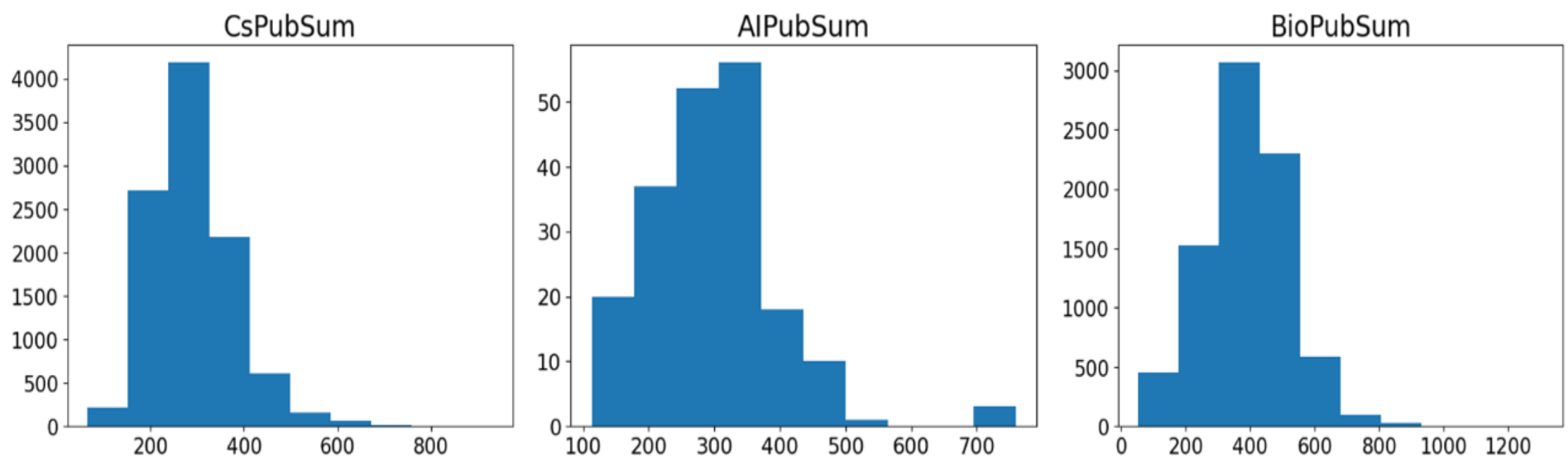

**Figure 4. The token distribution of demonstration in different datasets**

## 3.3 Language Model Call

The current study utilized five language models for highlight generation, including commonly used pre-trained language models such as GPT-2 and T5, as well as the recently popular ChatGPT model in the field of artificial intelligence. Previous prompt-based learning research has often employed the BERT model for experiments. However, given that the task in this study is text generation, and BERT, being a Transformer Encoder-based model, is not well-suited for generating long passages of text, we did not consider the BERT model in this research. We provide a brief introduction to the principles of the models used as follows:

**GPT2：** GPT2 (Radford et al., 2019) is a language model that slightly changes the structure of the original GPT and is trained on the large-scale data set WebText built by the research team. The most considerable GPT2 model parameter amount can reach 1.5B. GPT2 can handle zero-shot scenarios, is comparable to the supervised baseline model in reading comprehension tasks, and achieves SOTA on seven language modeling data sets.

**T5：** T5 (Raffel et al., 2020) is a Transformer-based pre-trained language model proposed by Google in 2020, achieving SOTA on a series of natural language processing tasks. Regarding model structure, T5 is almost the same as the original Transformer. Its outstanding feature is to convert all-natural language processing problems into text-to-text form so that a unified framework can be used to solve downstream tasks.

**Flan-T5：** Flan-T5 (Chung et al., 2022) is a T5 model after Instruction tuning. This model has powerful zero-shot, few-shot, and other capabilities, and its performance on many tasks exceeds that of the original T5 model. Instruction tuning is a training method for fine-tuning the model on a series of natural language processing tasks described by instructions, thereby improving the performance of the language model on unseen tasks.

**BART：** BART (Lewis et al., 2020) uses the standard Transformer architecture and absorbs the advantages of the two pre-training models, BERT and GPT, in the pre-training process. It can handle text generation tasks effectively and perform well in reading comprehension tasks.

**ChatGPT：** ChatGPT is a conversation model developed by OpenAI in 2022. It is a sibling model to InstructGPT(Ouyang et al., 2022), and both use *reinforcement learning from human feedback*(RLHF) algorithm for training. Its impressive ability to learn from small samples makes it an effective tool for natural language processing applications. Its high-quality text generation capability enhances content creation and summarization (Afjal, 2023).

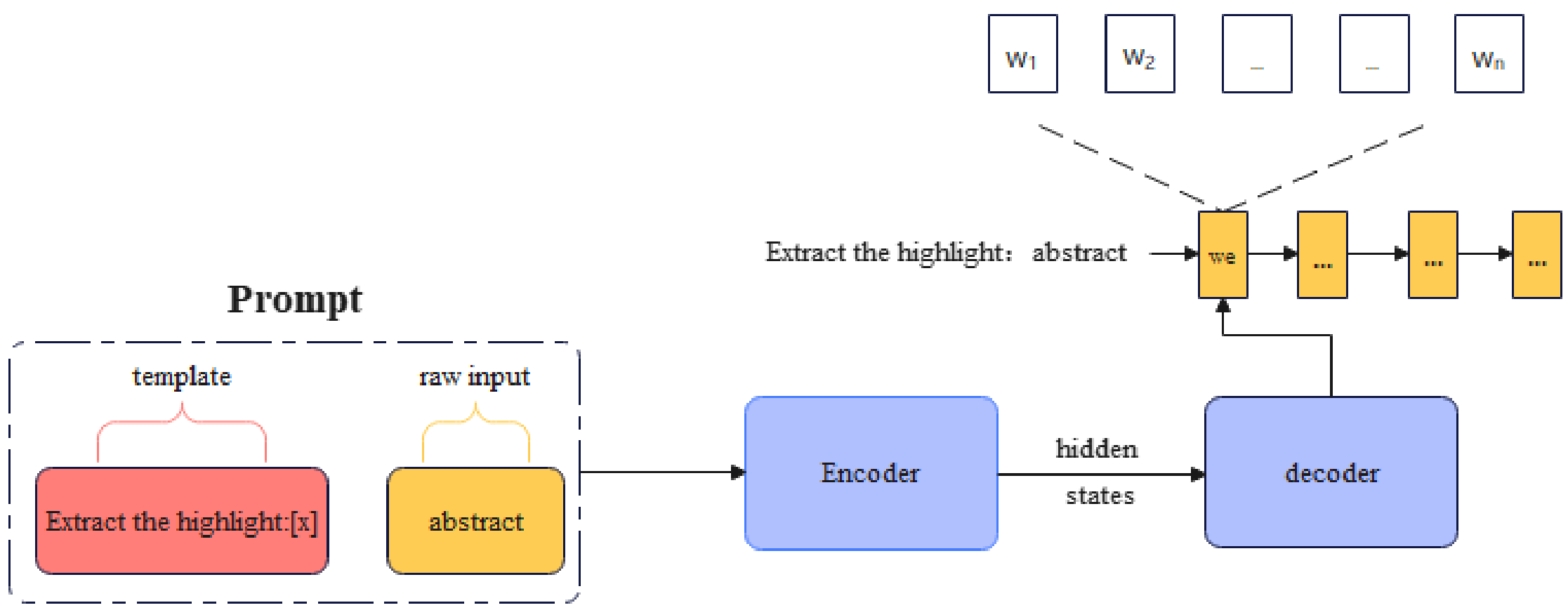

**Figure 5 The process of pretrained language model inference in local**

For the first four models, the model inference process is shown in Figure 5. We utilized the Transformers[2] library to build the language model and subsequently construct a prompt by filling the template with the input text. During the generation process in the decoder, the model generates tokens one by one. In each time step of generation, the model has n possible generated results, where n corresponds to the vocabulary size in the current pre-trained language model. In step $k$, the generated probability of each candidate word $w_j$ can be calculated using conditional probability $p(w_j|x_{1\sim k-1})$， where $x_{1\sim k-1}$ corresponds to the generated content before step $k$. Typically, we would choose $w_j$ with the highest probability as the generated result for that step. However, this approach may prioritize everyday vocabulary over specialized terms due to its reliance on general language corpora. Since our research corpus consists of academic papers, this method is unsuitable. Instead, we have opted for random sampling by selecting one word randomly from $q$ words whose generation probabilities exceed $p$ in our model. Here, $q$ and $p$ are hyperparameters.

[2] https://huggingface.co/

For ChatGPT, we utilize the API provided by OpenAI and submit prompts to the gpt3.5-turbo-0613 (a version of ChatGPT) model through the langchain[3] framework to obtain generated responses from the ChatGPT model.

# 4 Result

This chapter first introduced the evaluation metrics. In Section 4.2, we compared the generation performance of pre-trained language models with and without prompt-based learning and compared the best-performing model with the state-of-the-art (SOTA). The impact of prompt content and demonstration on model performance was further discussed in Sections 4.3 and 4.4, respectively.

## 4.1 Evaluation setting

We use the Rouge metric to measure the difference between generated and accurate content. The Rouge series of indicators was proposed by Lin in 2004 (Lin, 2014), including ROUGE-N, ROUGE-L, ROUGE-W, and ROUGE-S. The first two groups of indicators are widely used in tasks such as automatic summarization, so we choose ROUGE-N and ROUGE-L to evaluate the model effect. The indicators calculation formula is as follows:

$$ROUGE\ N(P) = \frac{N - gram_{match}}{N - gram_{prediction}} \tag{1}$$

$$ROUGE\ N(R) = \frac{N - gram_{match}}{N - gram_{reference}} \tag{2}$$

$$ROUGE\ N(F_1) = \frac{2 * ROUGE\ N(P) * ROUGE\ N(R)}{ROUGE\ N(P) + ROUGE\ N(R)} \tag{3}$$

$$ROUGE\ L(P) = \frac{LCS(prediction, reference)}{L_{prediction}} \tag{4}$$

$$ROUGE\ L(R) = \frac{LCS(prediction, reference)}{L_{reference}} \tag{5}$$

$$ROUGE\ L(F_1) = \frac{2 * ROUGE\ L(P) * ROUGE\ L(R)}{ROUGE\ L(P) + ROUGE\ L\ (R)} \tag{6}$$

N-$gram_{prediction}$ and N-$gram_{reference}$ represent the number of n-grams on the predicted and actual samples, and *N-gram*$_{match}$ denotes the number of matching n-gram terms between the predicted and actual samples. When calculating *ROUGE-N*, we ultimately use $F_1$ as a metric, which is formula (3). *LCS* (prediction, reference) represents the length of the longest common substring between the actual sample and the predicted sample. $L_{prediction}$ and $L_{reference}$ denote the predicted and accurate samples' lengths, respectively. When calculating *ROUGE-L*, we use $F_1$ as a metric, as given by formula (6).

## 4.2 Model performance on highlight generation

Table 3 compares the language models' performance on the three datasets. We categorized the

[3] https://python.langchain.com

models based on whether they utilize prompts. The "no prompt" condition is when we only input the abstract and obtain the model's inference result. The "add prompt" condition corresponds to when we used the prompts described in Section 3.2. The "add Demonstration" component includes examples within the template provided to the ChatGPT model, as shown in Figure 3. The results shown in Table 3 represent the best performance among the nine prompts for each model. The BART model performs the best without any prompting, though it has significantly fewer parameters and a smaller training corpus than ChatGPT, demonstrating its strong generation capabilities. After adding prompts, the performance of the ChatGPT model is improved dramatically, with scores far exceeding those of other models on various datasets and surpassing the existing state-of-the-art (SOTA) on the AIPubSum and BioPubSum datasets. Furthermore, by including demonstrations in the prompts, the model's generation performance was optimized on two datasets, widening the gap with the SOTA. This demonstrates the effectiveness of the approach proposed in this work。

An intriguing observation emerges from analyzing performance changes before and after incorporating prompts. For the GPT2 and BART models, prompts do not appear to enhance performance. In contrast, prompt-based learning can improve the performance of the T5 model, albeit with only a slight improvement in evaluation metrics such as ROUGE-1. However, for the Flan-T5 and ChatGPT models, the use of prompts has significantly impacted the inference results. To illustrate, in the case of Flan-T5, when only the abstract is used as input, its ROUGE-1 value on the CSPubSum dataset is merely 21.38; however, after adding a prompt, this value increases to 30.41 – a notable increase of 9 percentage points, which is also observed across the other datasets. We hypothesize this is because the Flan-T5 model and ChatGPT have undergone instruction-based fine-tuning. Therefore, when the input includes prompts like those described in Table 2, the models can better apply the text modeling capabilities acquired during the fine-tuning stage to the current task.

**Table 3. Model performance on the three datasets**

| Model | | CSPubSum | | | AIPubSum | | | BioPubSum | | |
|---|---|---|---|---|---|---|---|---|---|---|
| | | rouge1 | rouge2 | rougeL | rouge1 | rouge2 | rougeL | rouge1 | rouge2 | rougeL |
| no prompt | GPT2 | 24.19 | 10.44 | 20.49 | 22.66 | 8.72 | 18.65 | 22.11 | 10.01 | 18.53 |
| | T5 | 27.22 | 7.59 | 22.67 | 25.37 | 6.25 | 20.61 | 25.62 | 6.41 | 21.00 |
| | Flan-T5 | 21.38 | 6.69 | 18.92 | 19.69 | 5.76 | 17.45 | 22.52 | 7.59 | 19.94 |
| | BART | **33.51** | **12.17** | **28.44** | **29.66** | **8.39** | **24.29** | **30.48** | **9.95** | **24.91** |
| | ChatGPT | 25.99 | 10.12 | 23.72 | 24.62 | 8.75 | 22.27 | 25.05 | 10.26 | 22.51 |
| add prompt | GPT2 | 24.19 | 10.43 | 20.52 | 22.55 | 8.47 | 18.55 | 21.98 | 9.93 | 18.41 |
| | T5 | 30.08 | 10.03 | 25.88 | 26.93 | 7.03 | 22.49 | 27.37 | 7.74 | 22.71 |
| | Flan-T5 | 30.41 | 10.81 | 26.19 | 29.41 | 9.58 | 25.16 | 29.6 | 10.01 | 25.36 |
| | BART | 33.37 | 11.76 | 28.51 | 29.36 | 8.22 | 23.99 | 30.45 | 9.93 | 24.88 |
| | ChatGPT | **38.40** | **15.96** | **35.96** | **34.07** | **12.49** | **31.80** | **37.61** | **15.30** | **35.92** |
| add demonstration | | 38.89 | 15.77 | 36.08 | 38.99 | 16.33 | 36.69 | 39.2 | 15.6 | 36.34 |
| SOTA | | 39.9 | 19.2 | 37.2 | 33.8 | 12.7 | 30.8 | 35.5 | 13.7 | 32.4 |

## 4.3 The impact of different prompts on Model generation performance

In this section, we analyzed the performance changes induced by the different prompts on the models. Given that the ChatGPT model demonstrated the best performance in the previous experiment, we focus the subsequent discussion on its performance changes.

Table 4 shows the generation performance of the ChatGPT model when different prompt templates are provided without any additional demonstration. The prompt numbers in the first column of the table correspond to the order in which they are listed in Table 2. From the table, we can see that *prompt 8* has the overall best performance across the datasets. In the case of CSPubSum, where the ROUGE-1 scores for other prompts are generally below 36, *prompt 8* raised the score to 38.40. A similar trend is observed in the BioPubSum dataset. In contrast, using *prompt 6* as the input resulted in relatively poor generation performance for ChatGPT. On the CSPubSum dataset, the metric scores were even lower than those of the BART model without prompting (see Table 3). The other prompt templates, except for *prompt 7*, also performed moderately.

**Table 4. The performance of ChatGPT when using different prompt**

| Type of prompt | CSPubSum | | | AIPubSum | | | BioPubSum | | |
|---|---|---|---|---|---|---|---|---|---|
| | rouge1 | rouge2 | rougeL | rouge1 | rouge2 | rougeL | rouge1 | rouge2 | rougeL |
| *prompt 1* | 34.89 | 12.63 | 29.62 | 34.03 | 11.64 | 27.92 | 33.47 | 12.14 | 28.04 |
| *prompt 2* | 35.80 | 12.86 | 30.32 | 33.78 | 11.75 | 28.36 | 34.93 | 12.81 | 29.43 |
| *prompt 3* | 34.37 | 12.70 | 29.25 | 32.78 | 11.48 | 27.34 | 32.28 | 12.13 | 27.22 |
| *prompt 4* | 34.06 | 13.51 | 29.79 | 33.80 | 12.52 | 28.62 | 34.05 | 13.24 | 29.31 |
| *prompt 5* | 34.08 | 13.31 | 29.24 | 34.24 | 12.58 | 28.55 | 32.55 | 12.41 | 27.69 |
| *prompt 6* | 33.1 | 13.01 | 30.85 | 30.36 | 10.83 | 28.17 | 32.49 | 12.81 | 29.86 |
| *prompt 7* | 36.64 | 13.65 | 31.54 | **34.25** | **12.75** | 29.29 | 34.60 | 12.78 | 29.50 |
| *prompt 8* | **38.40** | **15.96** | **35.96** | 34.07 | 12.49 | **31.80** | **37.61** | **15.30** | **35.92** |
| *prompt 9* | 35.35 | 15.03 | 33.55 | 32.64 | 12.27 | 30.92 | 34.04 | 14.77 | 32.11 |

**Note**: The numbers in bold in the table are the highest indicator scores in the current column. Suppose there is no special instruction, the same as below.

The results in Table 4 suggest that the prompt template's content can significantly impact the model's performance. However, in the current experiment, we cannot definitively determine which specific information included in the prompt template leads to better generation performance. For example, *prompt 6* constrained the number of highlights generated, whereas *prompt 7* did not. The model's performance on the CSPubSum and AIPubSum datasets was much lower for *prompt 6* than *prompt 7*, indicating that including such a constraint in the prompt may not be beneficial. However, the comparison between *prompt 8* and *prompt 9* shows the opposite trend. *prompt 8* included a constraint on the number of highlights, which achieved better results than *prompt 9*.

The contrasting conclusions drawn from these two comparisons suggest that the impact of prompt design choices is more complex. As X. Liu et al. (2021) mentioned in their research, minor variations in the wording of prompts can significantly affect experimental results, even though their example was based on the BERT model. This phenomenon also exists in large language models like ChatGPT. Given the sensitivity of prompt design and the impracticality of manually identifying an optimal prompt template, alternative approaches may be necessary. For models like BERT and GPT-2, P-tuning can be used to search the prompt space and optimize the prompt design. However, optimizing the prompt design through model training is not feasible for models like ChatGPT, which are not open-source and have enormous parameters. Instead, researchers often rely more on expert knowledge and comparative performance evaluation to select prompts that exhibit better results.

### 4.4 The impact of demonstration on model generation performance

In Section 3.2, we described how adding a demonstration component to the prompt template provided to ChatGPT improves the model's performance. In this section, we further investigate the impact of the quantity and content of the samples included in the prompt template on the model's performance.

#### 4.4.1 The impact of examples quantity

Figure 6 shows the ROUGE-1 performance of the ChatGPT model on the three datasets when the number of training samples (k-shot) in the demonstration is increased. The overall trend indicates that as the number of demonstration examples increases, the model's generation performance improves accordingly. However, there are also some anomalous cases, such as in the AIPubSum dataset, where the model exhibited the best performance in the 1-shot scenario. One possible explanation for the anomalous performance on the AIPubSum dataset is that the training and test samples in this dataset differ significantly in content. Providing more demonstration examples may have caused the model to be overly influenced by the content in the demonstration, leading to bias when making inferences on test samples that deviate substantially from the demonstration content.

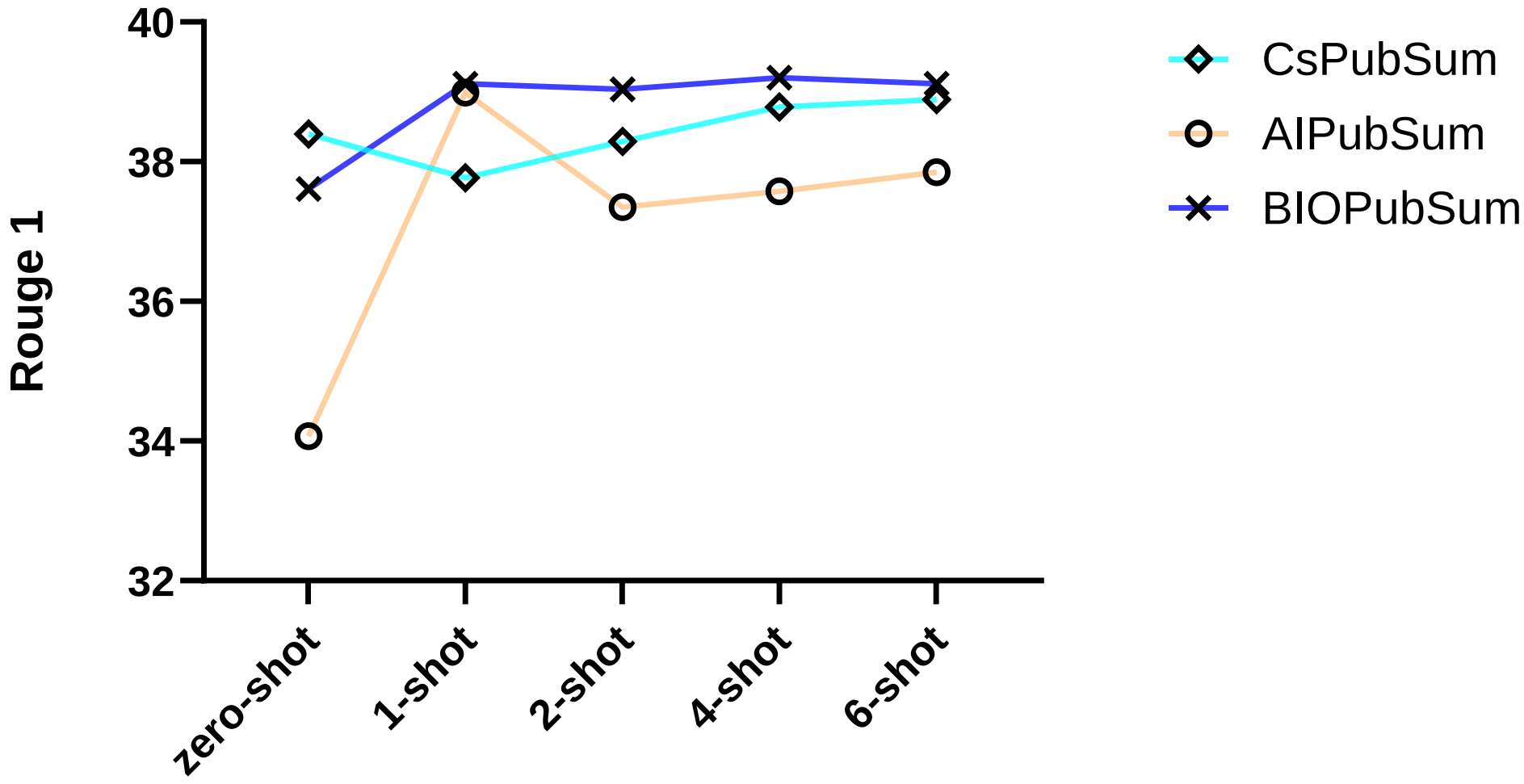

**Figure 6 Model Inference Performance on CSPubSum with Varying Demonstration Sample Sizes**

#### 4.4.2 The impact of examples content

J. Liu et al. (2021) observed in their experiments that when using GPT-3 for downstream tasks like text classification and question answering via In-context Learning, the content of the samples included in the prompt can significantly impact the results. When the number of samples in the prompt is fixed, selecting training samples that are more semantically similar to the test samples can lead to more significant performance improvements for the model. This study also observed the impact of different demonstration content on the model's performance. Table 5 shows the model's performance on the AIPubSum dataset when randomly selected training samples are included in the prompt. The metric scores exhibit significant variations, with the ROUGE-1 score ranging from as low as 36.08% to as high as 38.99%.

To address the high uncertainty associated with randomly selecting samples, we have adopted the approach proposed by J. Liu et al. (2021) in their research to identify the most suitable samples from the training set to include in the prompt template. For the abstracts in the test set for which Highlights need to be generated, we first compute the relevance between the current abstract and all abstracts in the training set. This study employs two approaches to calculate the relevance: 1) Utilizing the text-embedding-ada-002 model to represent the raw text as high-dimensional semantic vectors and then computing the semantic similarity; 2) Calculating the degree of N-gram vocabulary overlap between the text pairs. Then, we select the k most relevant abstracts from the training set and use their corresponding samples as the demonstration examples to be added to the Prompt. Finally, as shown in Figure 3, the Prompt is constructed and passed to the model for inference. We compared the model results obtained using these two relevance-based methods against those from randomly selecting demonstration samples, as shown in Table 6. To maintain clarity in the table, we only present the experimental results on the CSPubSum dataset. For the random selection approach, we conducted four experiments for each sample size scenario, ensuring that the samples included in the Prompt were unique in each run. We then took the average of the four experimental runs as the final result.

**Table 5 Variations in Model Performance with Different Samples Included in the Prompt**

| **Experimental rounds** | **AIPubSum（1-shot）** | | |
|---|---|---|---|
| | rouge1 | rouge2 | rougeL |
| round 1 | 38.99 | 16.33 | 36.69 |
| round 2 | 36.80 | 13.78 | 34.49 |
| round 3 | 36.88 | 13.93 | 34.58 |
| round 4 | 36.08 | 13.89 | 34.14 |
| Average | 37.17 | 14.51 | 34.97 |

**Table 6 Model Generation Performance with Different Sample Selection Strategies**

| **K-shot setting** | **Random** | | | **Embedding** | | | **N-gram** | | |
|---|---|---|---|---|---|---|---|---|---|
| | rouge1 | rouge2 | rougeL | rouge1 | rouge2 | rougeL | rouge1 | rouge2 | rougeL |
| 1-shot | 37.17 | **14.51** | 34.97 | 36.04 | 13.72 | 34.04 | **37.61** | 13.95 | **34.99** |
| 2-shot | 36.85 | 14.17 | 34.62 | 35.95 | 13.86 | 33.78 | **38.24** | **15.1** | **35.71** |
| 4-shot | **37.20** | 14.57 | **34.92** | 36.32 | **14.59** | 34.33 | 36.98 | 14.33 | 34.67 |
| 6-shot | **37.70** | **14.49** | 35.46 | 37.68 | 14.7 | **35.65** | 37.18 | 13.92 | 34.64 |

Compared to the three approaches, selecting samples based on the similarity of embedding-based vector representations performed poorly. In contrast, the more straightforward approach of using N-gram vocabulary overlap exhibited better results. This suggests surface-level textual information and features may be more critical than deeper semantic information when selecting demonstration samples. Further in-depth research would be needed to explain this observed phenomenon theoretically. Compared to the random selection approach, the N-gram-based method performs better in the 1-shot and 2-shot scenarios. Although the performance of the N-gram-based method began to decline as the sample size increased, considering the potential for random selection to sometimes choose samples that lead to poor performance, we recommend using the N-gram-based approach to select the demonstration samples in the Prompt. This can help ensure more stable output results from the model.

## 5 Case study

In this section, we analyze the extent to which the model-generated Highlights can replace those written by humans through specific examples. We randomly selected the abstracts and highlights of two papers from the dataset and had ChatGPT automatically generate highlights based on the given prompt template. We then compared the content-level differences between the two types of highlights, and the results are shown in Figure 7. Each row in Figure 7 represents the Highlights obtained through the two methods for the same paper. The blue portions in the "human-written" highlights and the yellow portions in the "model-generated" highlights indicate the overlapping content between the two types of highlights. For a given paper, if the blue portion in the "human-written" highlights takes up a more significant proportion, it suggests that the model has a higher recall rate and the generated results can sufficiently cover the key points emphasized by the human authors. Conversely, suppose the yellow portion in the "model-generated" highlights takes up a larger proportion. In that case, it indicates that the model-generated results contain less irrelevant information.

| **human-written** | **model-generate** |
| --- | --- |
| *We proposed a simultaneous evaluation method for multiple muscle loads by applying the optimization methodology.*<br>A satisfaction level function was derived to reveal the relationship between the muscle load and the subjective evaluation.<br>*The proposed method was applied to the design problem of handrail position for standing.*<br>*The proposed method can provide both objective and subjective information, and the method may be a great help to designers.* | *An evaluation of multiple muscle loads and a procedure for determining optimum solutions to ergonomic design problems is proposed.*<br>The simultaneous muscle load evaluation is formulated as a multi-objective optimization problem.<br>Optimum solutions are obtained for each participant, as well as a compromise solution for all participants.<br>*The proposed method provides both objective and subjective information to support decision-making.*<br>*The method was applied to the problem of designing the handrail position for the sit-to-stand movement.* |
| *We carried out a set of pilot guidelines for the design/evaluation of warning signs for young children.*<br>*The evaluation involved interviews and focus groups with parents, teachers, human factors experts, and other groups.*<br>*The findings offered broad support for the guidelines and some revisions.*<br>*The paper concludes with a set of recommendations for future work (e.g., the need for behavioral testing).* | *We conducted a study to refine guidelines for the design and evaluation of warning signs for young children.*<br>*The study involved interviews and focus groups with parents, teachers, human factors experts, and other groups.*<br>*The findings provided support for the guidelines and identified areas for improvement,* such as considering the target audience in more detail and addressing gender and special needs.<br>Similar findings were obtained for the evaluation guidelines, suggesting the need for additional methods and activities for testing signs.<br>We propose a revised set of guidelines and suggestions for a more comprehensive approach to sign design and evaluation.<br>*Future research topics include behavioral testing with children and their caregivers.* |

*Note: The colored portion indicates the semantic content overlap between human-written and machine-generated highlights.*

**Figure 7 The Highlights comparison of human written and model generate.**

Based on the results shown in Figure 7, the model demonstrates a more robust performance in terms of recall. Although there are differences in the word choices and expressions compared to the human-written highlights, the model-generated content is generally consistent in the key points covered. However, the model's precision is lower, as the generated results contain significant content outside the actual highlights. This suggests that further filtering of the generated content would be necessary for practical application, such as calculating the semantic similarity between each generated sentence and the original abstract and retaining the more relevant text. Overall, the method proposed in this paper can generate relatively high-quality highlights. This can assist authors in providing references when writing the highlights for their papers. On the other hand, it can automatically generate highlights for many papers that do not have existing highlights, thereby providing data assurance for large-scale text-based analyses that rely on highlights, such as bibliometric studies.

# 6 Discussion

Most researchers spend most of their time reading academic papers. Reviewing these publications can help us stay abreast of the latest developments and emerging methodologies in our respective fields. However, human beings' cognitive capacity is inherently finite. Given the sheer volume of literature available, it is not feasible to meticulously read and summarize every research paper. The prevalent approach for researchers is to conduct searches on journal websites such as Emerald, Elsevier, and others, using keywords, author information, and related criteria to compile a set of candidate papers, then leverage the title, abstract, and other metadata provided on these online platforms to develop a general understanding of the paper's content, which informs our decision on whether the publication merits further in-depth review by downloading the full text. In this process, for the paper's authors, clearly presenting the study's key findings and innovative methodologies to potential readers can significantly boost the paper's visibility and likelihood of being cited and downloaded. From the reader's perspective, gleaning valuable information about the research methods, significance, and key takeaways from succinct paper content can significantly reduce the burden of reading and then enhance overall research efficiency. This study focuses on the "Highlight" as a core component of academic papers, which has the potential to address the needs of both paper authors and readers, thereby fostering the positive development of research activities.

Prompt learning has been widely applied in the BERT model(Zhu et al., 2023). While the BERT architecture, which is founded on an encoder structure, is not primarily designed to address text generation problems, researchers can leverage prompt learning templates to reformulate tasks such as text classification into cloze-style fill-in-the-blank problems that align closely with BERT's pre-training objective. This alignment allows BERT to leverage the knowledge and representations acquired during the pretraining stage more effectively. With the emergence of models like Flan-T5 and ChatGPT, the effectiveness of prompt learning has become even more pronounced. By providing large language models with appropriate prompts and a few demonstrative examples, researchers have found that these models can achieve performance on par with traditional supervised learning approaches without updating the model parameters. To a large extent, the remarkable performance of these language models can be attributed to their massive parameter counts and the vast training corpora, which endows them with powerful modeling capabilities across a wide range of downstream tasks. However, as the analysis in Section 4.3 demonstrates, the prompts significantly impact the model's performance. In this study, we have manually crafted prompt templates, and recent research has also explored the use of deep learning models to derive optimal prompt representations for specific tasks automatically. These efforts have primarily focused on constructing prompts for models like GPT-2(Li & Liang, 2021). Regarding large language models like ChatGPT and Claude, the prevailing approach still relies on expert experience and empirical experimentation to select the most suitable prompt templates.

Yang(Yang, 2016) analyzed the highlight sections of 240 research papers across different academic disciplines. The findings reveal that authors from various fields tend to emphasize distinct aspects when describing the focus of their work. Wang(Wang et al., 2023) conducted a comparative analysis of the Highlight content in papers from the library and information science (LIS) field versus the computer science domain. In the LIS literature, the Highlight statements focus more on describing the implications of the reported studies. At the same time, computer science papers emphasize outlining novel methodologies and algorithms. These studies indicate that Highlights exhibit significant variations in text length and content descriptions across different academic disciplines. This could lead to highlight extraction or generation models trained on a single dataset that may not be generalized well enough to capture the highlights of papers in other research fields. In previous work on highlight abstraction and generation, scholars have often chosen to train multiple models on diverse datasets to address this challenge. However, this approach incurs increased computational and data collection costs, hindering wider adoption. The findings of this study demonstrate that by providing large language models with appropriate prompt templates, they can effectively extract high-quality Highlights from text across diverse domains. Furthermore, the model's performance can be further improved by incorporating domain-specific training examples into the prompts.

# 7 Conclusion and Future Works

In this research study, we leverage prompt learning methods to design corresponding prompt templates based on the current task. We then use the prompt templates and paper abstracts as input to the language model to generate Highlights. By comparing the model performance before and after adding the prompt templates, we observe a significant improvement in the generation performance of models that have undergone instructional fine-tuning, such as Flan-T5 and ChatGPT. Even without providing training samples, the model's performance can reach the state-of-the-art (SOTA) level of supervised training models. Furthermore, when adding a small number of training samples, as examples in the prompt template provided to ChatGPT, the model's performance is further enhanced. In our subsequent analysis, we examined the prompt template content and the number of example samples included in the prompts. We found that minor variations in the wording within the templates can significantly impact the results. In most cases, the model's performance positively correlates with the increase in example samples in the prompts. In the case study section, we compared the Highlights generated by our model against those written by the paper's authors using specific examples. The results indicate that the Highlights generated by our method effectively capture the author's intended research focus and critical contributions, which can facilitate knowledge sharing and dissemination.

This study has several limitations that warrant acknowledgment. Firstly, as discussed in Section 4.3, we observed that even minor changes in the prompt content can significantly impact model performance. However, due to the constraints of accessing the ChatGPT model via the API, we could not conduct a deeper analysis of this phenomenon at the model level. Secondly, our approach involved the manual creation of prompts based on expert knowledge, followed by experimental comparisons to determine the best-performing prompt for each model. This process presented several challenges, including the vast search space for manually formulated prompts, the time-consuming nature of identifying better-performing prompts, and the need to repeat this process for different models, as each requires a unique best-performing prompt. These factors rendered the experiments considerably cumbersome. Lastly, while the automatic generation of content shows promise in replacing manual work, there is still room for improvement, as the output generated by ChatGPT often includes irrelevant or redundant information.

The future work of this study can be approached through two primary avenues: 1) We plan to leverage automated methods to search for the most effective prompts for the Highlight generation task within a specific model. This approach will help enhance the quality of the model's output. 2) We intend to refine the content generated by the model further through adjustments to the prompt descriptions and the implementation of additional filtering techniques. For instance, if the model's output contains more than five Highlight descriptions, we can calculate the semantic similarity between them and eliminate sentences that are too alike, thus avoiding redundancy. Additionally, We can compare each Highlight description against the abstract and retain only those with higher semantic similarity to ensure the relevance of the generated content.

## Acknowledgments

This study is supported by the Open Funding Project of Laboratory of ISTIC-Springer Nature Joint Laboratory for Open Science (Grant No. ISN23012).

## Ethics declarations

### Conflict of interest

The authors have no competing interests to declare relevant to this article's content.